\documentclass{article}
\usepackage[utf8]{inputenc}
\usepackage{amsfonts}
\usepackage{amssymb}
\usepackage{amsmath}
\usepackage{booktabs}
\usepackage{graphicx}
\usepackage{url}
\usepackage[most]{tcolorbox}
\newcommand{\ignore}[1]{}
\def\IR{\mathbb{R}}

\newcommand{\CL}{{\cal L}}

\newcommand{\be}{\begin{equation}}
\newcommand{\ee}{\end{equation}}
\newcommand{\ba}{\begin{array}}
\newcommand{\ea}{\end{array}}
\newcommand{\bea}{\begin{eqnarray}}
\newcommand{\eea}{\end{eqnarray}}
\usepackage{pstricks,pst-node,pst-text,pst-3d}
\definecolor{lightgray}{gray}{0.95}
\usepackage{color}
\definecolor{keywordcolor}{rgb}{0.5, 0.1, 0.1}   
\definecolor{tacticcolor}{rgb}{0.1, 0.2, 0.4}    
\definecolor{commentcolor}{rgb}{0.4, 0.4, 0.4}   
\definecolor{symbolcolor}{rgb}{0.0, 0.1, 0.6}    
\definecolor{sortcolor}{rgb}{0.1, 0.5, 0.1}      
\definecolor{attributecolor}{rgb}{0.7, 0.1, 0.1} 
\usepackage[colorlinks=true,linkcolor=blue,urlcolor=blue,citecolor=blue]{hyperref}
\usepackage{upgreek}
\usepackage{listings}
\def\boldclass{\bf\sf}
\def\P{{\boldclass P}}
\def\NP{{\boldclass NP}}

\def\TC{{\boldclass TC}}
\title{ Large Language Models } 
\date{July 2023}

\author{Michael R. Douglas \\
    CMSA, Harvard University \\
    Dept. of Physics, Stony Brook University \\
    {\tt mdouglas@cmsa.fas.harvard.edu}
}

\begin{document}
\maketitle{}

\begin{abstract}
Artificial intelligence is making spectacular progress, and one of the best examples
is the development of large language models (LLMs) such as OpenAI’s GPT series.
In these lectures, written for readers with a background in mathematics or physics,
we give a brief history and survey of the state of the art, 
and describe the underlying transformer architecture in detail.
We then explore some current ideas on how LLMs work and how models trained to 
predict the next word in a text are able to perform other tasks displaying intelligence.
\end{abstract}
\vfill\eject

\section{ Introduction }
\label{s:intro}

At the end of November 2022, OpenAI released a system called ChatGPT which interacts with its users
in natural language. It can answer questions, engage in dialogs, translate between languages and write computer
code with a fluency and ability far exceeding all previous publicly available systems.  Although
it falls well short of human abilities in many ways, 
still the large language model technology of which it is an example
is widely considered to be a major advance in artificial intelligence.\footnote{
A few of the many other milestones in LLM development are
\cite{brown_language_2020,chowdhery_palm_2022,devlin_bert:_2018,radford_language_2019,raffel_exploring_2020}.}

Few developments in science and technology entered the popular consciousness as quickly as ChatGPT.
There is no mystery about why.  The ability to use language is a defining property of humanity, and for the first
time a computer is doing this well enough to make a comparison with humans interesting.  All of the hopes
and fears which have developed around AI, robots and technology more generally are being brought into the
discussion.  In my opinion this is justified; the speed of recent progress makes it urgent to better understand
AI, to forecast its capabilities and limitations, and to make wise decisions about its development and use.  
With great opportunities will come great challenges, which will concern all of us.

In these lecture notes we give an introduction to this subject for mathematicians,
physicists, and other scientists and readers who are mathematically knowledgeable but not necessarily
expert in machine learning or artificial intelligence.  
We begin with a very brief overview of AI in \S \ref{s:symcon} to explain some ideas we consider
to be essential context, the basic principles of the symbolic and connectionist approaches.
In \S \ref{s:lmhist} we define statistical language models and relate the history of transformer-based
LLMs up through GPT-4.  In \S \ref{s:pheno} we discuss measures of
what LLMs do and how well they do it.
We then give a precise explanation of simpler language models in \S \ref{s:details} and the transformer
architecture in \S \ref{s:trans}.  

It is amazing that a model defined by a few short equations, trained to go through a text and
simply predict each next word as it appears -- a task which seems only loosely related to any
definition of intelligence -- can do tasks which ``obviously'' require intelligence, such as 
solving word problems like the one in Figure \ref{fig:minerva} below.
At present nobody really understands how this works.
Even the interpretation of what LLMs are doing is controversial, ranging from the belief that
they are ``simply'' rearranging the sentences they 
were trained on, all the way to the belief that the LLMs are learning sophisticated models of the
world and that ``simply'' scaling up the computations will produce artificial general intelligence.
Any forecast for progress must take into account the current models' shortcomings --
lack of long term memory and ability to plan, tendency to make up facts and ``hallucinate,'' unreliability in logical reasoning, {\it etc.}.  Do these problems have technical solutions which will also look simple once we have them?
Or are they more significant barriers?

\begin{figure}[ht] \label{fig:minerva}

\begin{tcolorbox}[colback=blue!5!white,colframe=blue!75!black]
\begin{small}
\textbf{Question:}
Assume that the variance of the first $n$ natural numbers is 10, and the variance of the first $m$ even natural numbers is $16$. Compute $m+n$.
\end{small}

\tcblower
\vspace{-0.5mm}

\begin{small}
\textbf{Model output:}
Let $\sigma^2$ be the variance of the first $m$ even natural numbers, and $\tau^2$ be the variance of the first $n$ natural numbers. Then, we have $\tau^2=\frac{n^2-1}{12}$ and $\sigma^2=\frac{m^2-1}{3}$. We solve the system of equations: \begin{align*}
\frac{n^2-1}{12} &= 10\\
\frac{m^2-1}{3} &= 16.
\end{align*} This translates into the system \begin{align*}
n^2 &= 121\\
m^2 &= 48+1 = 49.
\end{align*} Therefore, $n = \boxed{11}$ and $m=\boxed{7}$, so $n+m = \boxed{18}$.
\vspace{1mm}
\end{small}
\end{tcolorbox}
\caption{ A question-answer pair solved by Minerva.  From Lewkowycz {\it et al} \cite{lewkowycz_solving_2022,CCBY}, 
``Solving quantitative reasoning problems with language models,'' 2022.}
\end{figure}

Most current work on LLMs takes an engineering and problem solving perspective, but there are many interesting
works which focus more on understanding how LLMs work.   One would think this should be far easier than
understanding how human brains work, as we have full knowledge of an LLM's microscopic workings
and can do a wide variety of experiments on it.\footnote{
At least, this was the case before March 2023.  Currently the weights and even the design parameters of GPT-4, 
the most advanced LLM, are held in confidence by OpenAI as proprietary information.}
These efforts are in their early days, but in \S \ref{s:study} 
we survey current approaches to understanding how LLMs do what they do.
We conclude in \S \ref{s:end} with more general discussion, some questions and 
potentially important developments to watch for.\footnote{
Other general reviews of LLMs include \cite{wolfram2023chatgpt,zhao_survey_2023}.}

Before we start, let me say a little about my own background.  I was trained as a theoretical physicist
and most of my contributions to science are in string theory and its interface with mathematics, but I have followed
AI fairly closely since the 80's and in detail since 2016.
In addition I spent eight years in quantitative finance where
I gained a good deal of ``hands-on'' experience with machine learning.  I have given many lectures telling computer
scientists about physics and physicists about computational topics, and benefited from conversations
with many people -- more than I can name here, but let me thank
Gerald Jay Sussman, David McAllester, Yann LeCun, Sanjeev Arora, Surya Ganguli, 
Jeremy Avigad, Vijay Balasubramanian, Dmitri Chklovskii, 
David Donoho, Steve Skiena,
Christian Szegedy, Misha Tsodyks, 
Tony Wu and the many speakers in the CMSA New Technologies seminar series,\footnote{
\tt https://live-hu-cmsa-222.pantheonsite.io/}
and Josef Urban and the AITP community.\footnote{\tt http://aitp-conference.org/2023/}
Thanks to David McAllester and Sergiy Verstyuk for comments on the first draft.
These notes would not have been possible without their
input and advice, and I hope their signal to noise ratio approaches that of what they shared with me.

\section{ Symbolic and connectionist AI }
\label{s:symcon}

The goal of artificial intelligence is to build computational systems which can perform tasks
which require intelligence.  Although intelligence is hard to define precisely, an operational
definition suitable for LLMs is ability at tasks requiring language, reasoning, and planning, as judged by 
humans who interact with the system.  Some famous and difficult ``challenge tasks'' include playing chess \cite{shannon1950xxii},
proving mathematical theorems \cite{newell1957empirical}, and answering natural language questions 
using generally known facts and common sense.%
\footnote{This challenge originates with Turing's famous test, but the restriction to question answering
makes it better defined and testable using benchmarks, standardized question-answer sets.
Discussion of the original test can be found at
{\tt https://en.wikipedia.org/wiki/Turing{\_}test}}

These problems have been the subject of intense investigation since the mid-50's, and a few
textbooks and histories are \cite{mccorduck2004machines,nilsson2009quest,russell2010artificial,sejnowski2018deep}.
Essentially from the start,
two broad approaches were set out, which would later be called symbolic and connectionist AI.\footnote{
Symbolic AI is sometimes called ``GOFAI'' for ``good old-fashioned AI.''
Related terms include ``rule based,'' ``logic based,'' ``expert system''
and ``feature engineering.''  
The connectionist approach has many other names,
reflecting its mixed ancestry: ``neural,'' ``deep learning,'' 
``parallel distributed processing,'' ``differentiable,'' and ``representation learning.'' 
}
The symbolic approach originated in mathematical logic and generative linguistic theory, and tracked
the development of computer technology (both hardware and software) as a tool for
solving practical and scientific problems.  Central topics in this approach are 
formal logic and language theory, search and heuristics, 
and engineering techniques for designing and building large and complex systems.

Symbolic AI systems are designed, meaning that their creators
develop a detailed understanding of the task, and encode this understanding into the system
by programming, by hardware design and otherwise.  As an example, consider the task of parsing: given an
input string of words, determine its grammatical structure.  Most of us learned to diagram sentences
in elementary school, and although linguists have developed far more sophisticated notions of grammar,
this simple notion gives the right idea.  The grammar of a language is encoded in rules, which belong to
a formal framework -- see the appendix for the example of context-free grammars.
Given such a framework, one can design a parsing
algorithm which takes as input a rule set and an input string, and produces an output which states whether the
string is a grammatical sentence and if so makes its structure explicit.  This is a symbolic AI approach, not 
because the words ``symbolize'' anything (after all grammar does not have to come with meaning),
but because the grammatical rules and the parsing algorithm (including its internal data structures)
have a clear meaning to their designers.

Symbolic methods have had considerable success at many tasks requiring intelligence, famously
including chess playing \cite{hsu2002behind} and symbolic algebra\footnote{\tt https://www.sigsam.org/}
as well as more prosaic but very central tasks
such as translating high level computer languages to machine code (compiling).
And a great deal of work has been done to broaden their scope, for example to build
question answering systems such as the well known IBM Watson.  Many valuable techniques came out of this;
ways to systematize rules into ``knowledge bases'' or ``knowledge graphs,'' methods for automated logical reasoning,
and so on.
But it was long ago realized that
once one goes beyond ``formal worlds'' such as chess and algebra to the complex and messy situations
of real life, although one can postulate rules which capture many truths and can be used for reasoning, rules
which are valid in all cases are very rare.   Furthermore,
the sheer number of rules required to cover even the likely possibilities
is very large.  These difficulties were addressed by implementing probabilistic reasoning 
and by getting teams of humans to develop the requisite
enormous rule sets, leading to the ``expert system'' approach which was applied (for example) to
medical question answering.  Cyc,\footnote{\tt https://cyc.com/} 
an early and well known expert system, is commercially available and has a database
of commonly known facts with over 25 million 
assertions; however this is dwarfed by knowledge bases such as Wikidata (over one billion
``facts'') but which do not have a systematic reasoning engine.  It is clear that any approach which depends on
careful human analysis of such large knowledge bases is impractical.


Meanwhile, a very different ``connectionist'' approach to AI was being championed by other 
researchers.  They drew their inspiration from hypotheses about how
the brain works, from information theory and statistics, from physics and other natural sciences,
and from applied mathematics and particularly optimization theory.
These diverse points of view
came together in the 1990's and led to a great deal of interdisciplinary work,\footnote{
I first learned about this from \cite{mackay2003information,mumford2002pattern}.}
of which the part most
related to AI and which lies behind LLMs is called machine learning (ML).

The usual starting point in modern treatments of ML is to rephrase a task as a statistical inference 
problem based on a large dataset.  A canonical example is image recognition -- say, given an array
of pixels (light intensity values), estimate the probability that the image depicts a cat.  Rather than
design a system to do this, one starts with a large dataset of images with labels (cat and non-cat).
One then designs a very general statistical model and ``trains'' it on this dataset to predict the
label given the image.  This is supervised learning, one can also do ``self-supervised'' learning in
which the system predicts some part of the data given other parts (say, filling in part of an image).
A third standard ML paradigm is reinforcement learning, which applies to tasks which involve
choosing actions to fulfill a longer term goal.  The classic example is game playing, as in AlphaGo
and AlphaZero.

In any case, since the problem is formulated statistically, it is possible to consider the training dataset
one item at a time, and use it to incrementally improve the model.
This is almost always done by formulating the task in terms of an ``objective function''
which measures the quality with which it is performed, for example the accuracy with which
correct labels are assigned to images.  One then takes a parameterized model and trains it by
optimizing this function, evaluated on the training dataset, 
with respect to the model parameters.  For the classic models of statistics
this can be done analytically, as in a least squares fit.  For more general models
one uses numerical methods, such as gradient descent.  Either way, a central question of statistics
and machine learning is generalization, meaning the extent to which the model well describes data not in the
training set but 
sampled from the same probability distribution.  
A well known principle which speaks to this
question is ``Occam's razor,'' that simpler models will generalize better.  This is often simplified to the rule that
a model should have the minimal number of parameters needed to fit the dataset.  

Not all machine learning systems are ``deep learning'' \cite{LeCun2015DeepL}
or ``connectionist'' \cite{Rumelhart1986AGF}.
These terms generally refer to the use of neural networks with 
large numbers of parameters which provide effective universal function approximators.
While the idea is very old \cite{rosenblatt1958perceptron}, before 2012 it was widely believed to be impractical.
One argument for this was the ``dull side'' of Occam's razor --
models with so many parameters were destined
to overfit and would not generalize.  Evidently this is not the case, leading to concepts such as ``benign overfitting.'' 
\cite{bartlett_benign_2020,belkin_understand_2018} 
Another argument was that the objective functions for these models are highly nonconvex and
optimization would get stuck at poor quality local minima.  This can be a problem, but turns out to
be solvable for reasons that are partially understood \cite{choromanska_loss_2015,goodfellow2016deep}.
Finally, despite the effectiveness of the trained model in performing a task, the large number of parameters often
makes it very hard to understand how such a model works, and why a given input produces a particular output.
This ``interpretability problem'' remains a key issue with deep learning models, and is the subject of much research
\cite{guidotti_survey_2019}.

There are many other variations and hybrid approaches in the story.  
Another important one is the ``pattern recognition''
approach \cite{bishop2006pattern,mumford2010pattern}.  
This is also based on statistical inference but -- like the symbolic approach -- it emphasizes
the value of detailed understanding of the problem domain in designing the system.
For example, one could hand-code the initial layers of an image recognition
network to detect lines or other significant ``features'' of the image.   But unlike a purely symbolic
approach, these features would be used as input to a general statistical or neural model.

Another concept which illustrates the relation between the two approaches
is probabilistic reasoning, the use of rules such
as ``the chance of rain when it is cloudy is 50\%''.
One can state and use such rules in a symbolic approach (see for example \cite{koller2009probabilistic}),
the essential distinction with connectionism
is not the use of probabilities but rather the representation of knowledge in terms of explicit 
and meaningful rules.

As we suspect every reader has already heard, the symbolic approach was dominant from the early days until
2012, and (along with many other successes) led to a superhuman chess player, but seemed inadequate
for our other two challenge tasks (theorem proving and question answering).
In 2012 the connectionist approach surpassed other approaches to computer vision \cite{krizhevsky2017imagenet}, 
and ever since neural
systems have gone from triumph to triumph.  In 2017 the deep learning system AlphaZero
surpassed the symbolic AI chess players (and of course humans).  
Over the last few years, transformer models trained on a large
corpus of natural language to predict each next word as it appears,
have revolutionized the field of natural language processing.  As we 
write this the state of the art GPT-4 demonstrates truly remarkable performance at question answering,
code generation and many other tasks \cite{bubeck_sparks_2023}.

The simplest and arguably deepest explanation for this history is that it is a consequence of the
exponential growth of computational power and training datasets, which continues
to the present day.  Given limited computing power and data, the ability of the symbolic 
and pattern recognition approaches to
directly incorporate human understanding into a system is a significant advantage.  On the other hand, 
given sufficiently large computing power and data, this advantage is nullified and may even become disadvantageous,
as the human effort required to code the system becomes the limiting resource.  This point, that the most
significant advances in AI (and computation more generally) have come from hardware improvements and
replacing human engineering with data-driven methods, is forcefully made by Sutton in his ``bitter lesson'' essay \cite{Sutton_2019}.  
In \S \ref{s:lmhist}, \S \ref{s:pheno} and \S \ref{s:end}
we will discuss scaling laws and evidence for and against the idea that by continuing along
the current path, training ever-larger models on ever-larger datasets, we will achieve AGI (artificial general
intelligence, whatever that means) and the realms beyond.

Up to now the symbolic and connectionist approaches have generally been considered
to be in tension.\footnote{
To better discuss this point one should refine the symbolic-connectionist dichotomy into multiple axes:
system design versus learning from data; meaningful rules versus uninterpreted models;
combinatorial versus differentiable optimization; deterministic versus probabilistic.
}  
There is another point of view which considers them complementary, with a symbolic approach
better suited for certain problems (for example logical reasoning) and connectionist for others (for example image
recognition).  Given this point of view one can seek a synthesis or ``neurosymbolic'' approach, advocated in 
many works \cite{bader_dimensions_2005,garcez_neurosymbolic_2020,lake_human-level_2015}.

But are they in conflict at all?  Another reconciliation is the hypothesis that problems which 
in the symbolic approach are solved using rules and algorithms, are also being solved
that way by neural systems and in particular by LLMs.  
However, rather than the algorithms and rules being coded by humans, as the result of its training procedure
the LLM has somehow learned them, encoded in its networks in some as yet mysterious way.
This vague hypothesis can be sharpened in many ways, in part by proposing specific mechanisms by which
algorithms and rules are encoded, in part by making general claims about the algorithms which are being learned.
We discuss these ideas in \S \ref{s:study} and \S \ref{s:end}.

\section{ Language models }
\label{s:lmhist}

Throughout the history of linguistics, languages have been described in terms of
rules: rules of grammar, phonology, morphology, and so on, along with logical and other frameworks for
describing meaning.
This remains the case in Chomskyan linguistics and in 
much of theoretical linguistics.

By contrast, LLMs are statistical language models, meaning that they
encode a probability distribution on strings of words, call this $P(w_1\ldots w_L)$,
which approximates the distribution realized by a large body (or ``corpus'') of text in the language.  
The simplest example is the frequency or ``1-gram'' model defined by taking the words to be
independently distributed, so
\be
P(w_1 \ldots w_L) = \prod_{i=1}^L P(w_i)
; \ \ %
P(w) = \frac{\mbox{number of occurrences of } w \mbox{ in the corpus} }{ \mbox{ total number of words in the corpus } }.
\ee
Of course, this model captures very little of the structure of language,
which involves dependencies between the word choices.

LLMs are generative models,\footnote{
Many different definitions of this term can be found in the literature. } 
by which we will mean that there is
a practical method for sampling from the distribution.  To explain this,
consider a word prediction task in which some words in a string are given (the ``input'') and others left blank
(the ``output'').  Given a probability distribution $P(w_1\ldots w_L)$, there is a corresponding
conditional probability distribution for the output given the input. 
As an example, suppose we are given the string ``The cat [BLANK] outside,''
where ``[BLANK]'' is a ``token'' which marks the position of the missing word.
The relevant conditional probabilities might be
\bea\nonumber
P( \mbox{the cat went outside} \;|\; \mbox{the cat [BLANK] outside}) &=& 0.5 \\ \nonumber
P( \mbox{the cat sat outside} \;|\; \mbox{the cat [BLANK] outside}) &=& 0.2
\eea
and so on, summing to total probability $1$.  
In the masked word prediction task, the model must determine (or sample from) this distribution.  

A particularly convenient case is to give the conditional probability of the word which follows a given string,
which we denote as
\be \label{eq:cond1}
P( w_{n+1} \; | \; w_1\; w_2\;\ldots\; w_{n-1}\; w_n ).
\ee
By sampling this distribution to get a new word $w_{n+1}$ and appending it to the end, the string can
be extended one word at a time.
Repeating this process gives an arbitrarily long string, which by the laws of probability is a sample from
the original probability distribution $P(w_1\ldots w_L)$, for example
\bea\nonumber
P( \mbox{the cat went outside} ) = 
 P( \mbox{the} ) P( \mbox{cat} \; | \; \mbox{the}  ) P( \mbox{went} \; | \;  \mbox{the cat} ) 
P( \mbox{outside} \; | \; \mbox{the cat went} ).
\eea
This factorization of the probability into successive conditional probabilities
defines the class of autoregressive models.  One could furthermore require that the
conditional probability Eq. \ref{eq:cond1} depends only on the $k$ most recent words, in which case
one would have a Markov model whose state is a string of $k$ words.

To evaluate how good a language model is, we want to quantify 
how well its probability distribution approximates that
of the corpus (the empirical distribution).  The standard measure of this is 
the cross entropy.
For an autoregressive model this is a sum of terms, one for
each word in the corpus,\footnote{ 
With the sign, ${\cal L}\ge 0$ and better models have smaller $\cal L$.  
The term ``loss function'' is often used for an objective function with these properties. } 
\be \label{eq:loss}
{\cal L} = -\frac{1}{N}\sum_{i=1}^{N-n} \log P( w_{i+n} \; | \; w_{i}\; w_{i+1}\;\ldots\; w_{i+n-1} )
\ee
One also refers to $\exp -\cal L$ as ``perplexity.''
In a machine learning approach, we can use Eq. \ref{eq:loss} as an objective function and minimize it as a function
of the network parameters to train the network.  We can then apply the many tools of ML: backpropagation,
splitting the sum into batches, varying the learning rate and so on, to get an efficient and effective model.
While the details are an art which depends on the particular 
domain and model architecture,\footnote{
In CS this term generally refers to the large scale arrangement of components of a system.}
conceptually these are much the same for LLMs as for other machine learning models.

This statistical approach to modeling language has been pursued since the late 80's \cite{brown1993mathematics,jurafsky2009speech,manning1999foundations}
and many models were developed,
such as the recurrent neural network (RNN)  which we will describe in \S \ref{s:details}.  Following the general
machine learning experience that supervised tasks (learning from input-output pairs) are easier than unsupervised
tasks, many of these works addressed machine translation and parsing, for which
there are good labeled datasets (documents with their translations; sentences with their grammatical structure).
However unlabeled datasets are much larger and by 2015 or so there was a sense that 
self-supervised learning was the next frontier \cite{LeCun_2015}, leading to more focus on masked word prediction.

The history of transformer models starts with the 2017 proposal of Vaswani {\it et al} \cite{vaswani_attention_2017}.  
Their model was designed for a translation task and was more complicated than what we will explain
in \S \ref{s:trans}, but the essential idea to use attention and positional encoding to represent all the relations
between the words in a text originated here and is fully present.  

The transformer architecture was taken up by many groups, and particularly influential 2018 works include
BERT \cite{devlin_bert:_2018} and GPT \cite{radford_improving_2018}.  BERT was trained by masking arbitrary words in a sentence (not just
the next word), which allows the model to look backward and forward for context and leads to better results.  
However it is not straightforward to sample from such a model, and eventually the simpler next word prediction
approach followed by GPT won out.

Both of these models, and most work of this period, followed the paradigm of pretraining followed by fine tuning.
The idea was to first train for word prediction on a very large corpus, to get a general purpose model.  This would
then be adapted to specific tasks such as question answering by fine tuning.  This
means doing a second pass of supervised learning on a much smaller labeled dataset, replacing next word prediction
by the objective function for the specific task.  Say we are doing question answering, 
this could be  the accuracy of the answers.
This two step procedure was justified by the notion of transfer learning, meaning that the capabilities of the general purpose
model ``transfer'' to related but different tasks.
This approach led to SOTA\footnote{
State of the art, in other words an improvement over all previously evaluated models.
}
results on many benchmarks and motivated much further work.  

Most importantly, a great deal of
ingenuity and hard work was put into solving the engineering problems of training larger and
larger models on larger and larger datasets.  As for the data, a lot of text is available on the web,
with one much used archive of this data provided by Common Crawl.\footnote{\tt https://commoncrawl.org/}
Training can largely be done in parallel by dividing up
this data, and the availability of large clusters of GPU-enabled servers at industrial labs and through cloud computing
meant that sufficient computing resources were available in principle.  However, the overall cost of training scales
as (at least) the product of model size and dataset size, and this was becoming expensive.  While the precise
cost figures for the GPT series are not public, it is estimated that a single training run of the largest GPT-3
models cost tens of millions of dollars.  To motivate and efficiently carry out such costly experiments, 
one needs some ability to predict
in advance how changes in model and dataset size will affect
the training methods (for example the optimal choice of learning rate) and performance.

\begin{figure}[ht]
\noindent \centering{} 
\includegraphics[width=\textwidth]{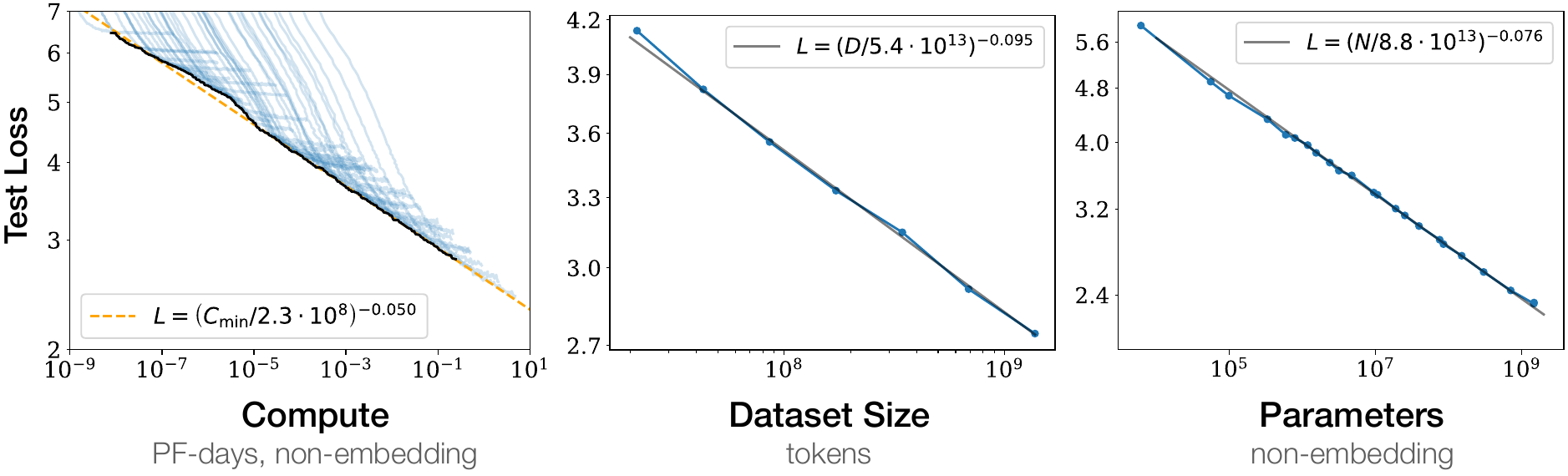}
\caption{Language modeling performance as a function of model size, dataset size, and amount of compute 
used for training. From Kaplan {\it et al}, ``Scaling Laws for Neural Language Models,'' 2020 \cite{kaplan_scaling_2020}.
\label{fig:BasicPowerLaws}}
\end{figure}

An important advance in this direction was the observation of power law scaling in language
model performance \cite{kaplan_scaling_2020}.  Figure \ref{fig:BasicPowerLaws} plots the test loss\footnote{
This is Eq. \ref{eq:loss} (minus log perplexity) evaluated on texts which were removed or ``held out'' of
the training set, to get a measure of generalization ability.
}
against the logarithms of the sizes and compute resources used, and these straight lines correspond to a
power law relation between size and perplexity.  This scaling holds over many decades in model
size and, while the exponents $\alpha \sim -0.076$ to $-0.095$ are rather small, this is a strong
argument that larger models will have better performance.
These ideas were also used to determine optimal model-dataset size tradeoff \cite{hoffmann_training_2022}
and the scaling of hyperparameters \cite{yang_tensor_2022}.
These results were a significant input into the decision to do this very expensive research.

\begin{table}[ht]
\centering
\begin{tabular}{lclc}
\toprule
Year & Model & Number of Parameters & Dataset size (tokens) \\
\midrule
2018 & GPT   & 110M                & 1B         \\
2018 & BERT & 340M & 3B \\
2019 & GPT-2 & 1.5B & 10B \\
2020 & GPT-3 & 175B & 500B \\
2022 & PaLM & 540B & 780B \\
2023 & GPT-4 & 1.4T (?) & ? \\
\bottomrule
\end{tabular}
\caption{Large Language Models (M/B/T = million/billion/trillion).  In many cases several model
sizes were considered; we quote the largest. }
\label{tab:model_parameters}
\end{table}

Now it should be realized that, while the measure being improved here is fairly objective, still there was
no strong reason to think that improving it would lead to models with qualitatively new ``emergent'' capabilities.
But it appears that this is what happened: GPT-3 and its fine-tuned cousins (such as Codex)
were able to do tasks, such as write computer code from a natural language description, for which smaller
models were almost worthless.\footnote{ A quantitative version of this claim is that performance for the
``emergent'' capability improves rapidly at some threshold value of the word prediction loss.
This claim is disputed, see
\cite{Schaeffer2023AreEA,wei_emergent_2022} for discussion. }
We will discuss more of this progress shortly, and 
speculate a bit in the conclusions.

One of the most interesting LLM phenomena is in-context learning,
first discussed in the original GPT-3 paper \cite{brown_language_2020}.  
This refers to the ability of an LLM to carry out tasks different from its original objective
without modifying its parameters, indeed without
any need for additional training on the new task (fine tuning).
Rather, after being given (as input text)
a few examples of input-output pairs, the LLM can be given another input and will
generate a suitable output.  Say the new task is question answering, then after a few question-answer
examples the LLM will answer the next question it is given.  While intuition based on human
abilities might find this unremarkable, it is actually quite unusual for an ML model
and this is why the pretraining-fine tuning paradigm was the usual approach in previous work.
Of course the training set already contains many examples of QA pairs.
More striking are tasks which are not much represented in
the training set, such as finding anagrams or rearranging letters in words.  
One can even do in-context ``meta-learning'' of machine learning tasks such
as linear regression (see \S \ref{s:pheno}).

Once it is established that the model can generalize from a few examples, a further step towards human
capabilities is to try zero examples, instead simply explaining the task in natural language.
At this point it becomes difficult to classify the tasks -- should we consider the task of writing code from
a natural language specification to be a form of translation, or an example of explaining the task, or something
else?  The relation between the input text or ``prompt'' and the output has many surprising features.  For example,
a standard technique in LLM question answering which measurably improves performance
is to precede the question with a prompt such as ``I will answer
this question helpfully and truthfully.''  Is this somehow biasing the network towards certain texts and away from
others (after all the internet corpus is hardly a reliable source of truth) ?  Suppose we have a theory of how this
works, how can we test it?  Does the model ``know'' anything about the truth of statements? 
\cite{burns_discovering_2022,li_inference-time_2023}

As has been much reported, one of the major difficulties in using LLMs for practical tasks is their propensity
to invent facts (especially citations) and their limited ability to do logical reasoning, algebra and other symbolic
tasks.  A device for improving this, called ``chain of thought prompting,'' is to give examples (say of question
answer task for definiteness) with some intermediate reasoning steps spelled out.  This was used in the Minerva
QA system \cite{lewkowycz_solving_2022} which produced the example in Figure \ref{fig:minerva}. 
Still the fraction of problems it solved correctly is around 50\% (the later GPT-4 is similar).
Even for simpler questions, the reliability of GPT-4 is more like 90\%.  
Much current research is devoted to this problem of reliable reasoning, as we discuss in \S \ref{s:how}.

\section{ Phenomenology of language models }
\label{s:pheno}

In this section we discuss general claims, ``non-invasive'' experiments,
and theoretical arguments which do not depend on ``microscopic details'' of the models
such as the trained weights.\footnote{
We are calling this ``phenomenology'' following the physics use of the term, 
not its use in psychology and philosophy to describe  the study of subjective experience.
}
This includes evaluation of model capabiliities, qualitative observations and scaling laws.

What can LLMs do?  There is a huge body of work on this question, and any attempt
to review it would rapidly go out of date, but let us review the primary method for
studying it.  This is benchmarking, the
development of standardized sets of test items for which model accuracy can be evaluated 
in a reproducible way.
This is in principle straightforward if the input corresponds to a single correct output,
as in multiple choice question answering.\footnote{
A potential pitfall is that 
after a benchmark is published, the test items can find their way into future training data
and then be solved by memorization.
Methods to detect and prevent this are discussed in the references.}
If the answer is free-form text, one can use
text comparison metrics such as the ROUGE score.  One current standard for evaluating LLMs,
BIG-bench \cite{srivastava_beyond_2022}, combines 204 language tasks (at first
publication; they accept new tasks) including translation, QA, puzzle solving, text classification
and summarization, and tests of common sense reasoning.   
A leaderboard listing the current best LLMs is at \footnote{
{\tt https://paperswithcode.com/dataset/big-bench}}.
Another is the EleutherAI ``Language Model Evaluation Harness''\footnote{
{\tt https://github.com/EleutherAI/lm-evaluation-harness}} and leaderboard.\footnote{
{\tt https://huggingface.co/spaces/HuggingFaceH4/open{\textunderscore}llm{\textunderscore}leaderboard}}
The benchmark suite HELM
\cite{liang_holistic_2022} measures additional metrics such as tendency
to repeat copyrighted material, bias, toxicity and the like.

Reasoning ability is of particular interest for mathematical and scientific applications -- of course
we all look forward to the day when computers will help us grade assignments, referee papers and
do our research.  There are many benchmarks for solving logical problems expressed in natural language.
Benchmarks for mathematical theorem proving
include NaturalProofs \cite{welleck_naturalproofs_2021}, MiniF2F \cite{zheng_minif2f_2022} and
ProofNet \cite{azerbayev_proofnet_2023}; as of mid-2023 LLMs (and the best
other systems) can find many proofs (20--80\%) but still fail on some seemingly
easy cases. 
Simpler aspects of reasoning which have benchmarks are the ability to deal with negation \cite{zhang_beyond_2023},
consistency (between different phrasings of the same question) \cite{jang_consistency_2023},
and compositionality (the ability to analyze statements and problems into simpler parts, solve these and
combine the results) \cite{press_measuring_2022}.

Natural language tasks are very complex, and benchmarks constructed from real world data
cannot be used directly in theoretical considerations.  For this purpose one generally defines ``toy worlds''
and generates synthetic data.  The possibilities are endless, but some which have been used are 
arithmetic problems (decimal arithmetic; modular arithmetic), game play, solving systems of equations,
and parsing formal languages.  A particularly interesting task is linear regression \cite{garg_what_2023};
since this is the prototypical case of statistical inference, a system which learns to do it can be said to be
``learning how to learn.''

Coming to scaling laws, denote the model size (number of parameters) as $P$ and the
dataset size (number of tokens in the corpus) as $D$, then there are two general regimes.
If we hold one of these (say $P$) fixed and take the other (say $D$) to infinity, then a law
of large numbers applies and $\CL \sim 1/D$.  On the other hand, if we take one parameter
very large and study the dependence on the other, nontrivial power law scaling can emerge.
In principle one can get different exponents for $D$ and $P$, suggesting the ansatz
\be\label{eq:scaling}
{\CL}(P,D) = \left[ \left(\frac{P_c}{P}\right)^{\alpha_P/\alpha_D} + \frac{D_c}{D} \right]^{\alpha_D} .
\ee
where $\CL$ is test loss Eq. \ref{eq:loss} computed in an optimally regularized model.\footnote{
Regularization is a standard technique in statistics and ML used to control overfitting by models with too many
parameters.  
If one does not regularize one sees other phenomena such as double descent \cite{belkin_reconciling_2019}.
For further discussion see \cite{bartlett_deep_2021,belkin_fit_2021}.
}
This is a good fit to Figure \ref{fig:BasicPowerLaws}.

While in Figure \ref{fig:BasicPowerLaws} the two exponents appear to differ, there is not
really convincing evidence that this is significant.
Before working hard on this, one should ask if there is any way to control the
many choices involved, so as to define universal exponents.
One context in which this can be studied systematically is transfer learning, by distinguishing
the dependence on the pretraining and fine tuning datasets \cite{hernandez_scaling_2021}.
Another relevant and practical question is whether one can prune the dataset to improve
the scaling.  It is intuitively plausible and can be shown in
examples that sets of data items are worth more if they are diverse than if they are similar.
The challenge is to find simple ways to quantify this similarity;
in \cite{sorscher_beyond_2022} many proposals are studied.

Scaling laws can arise in many ways, not specific to language models.  One hypothesis is that the data lies
on a low dimensional submanifold in a higher dimensional space.\footnote{In \S \ref{s:details} we explain
how text can be thought of embedded in a high dimensional space.}  Both the number of parameters 
and the number of points required to fit this manifold go as the dimension $d$ of the manifold, and
this leads to $\alpha_P=\alpha_D=4/d$ (the precise coefficient $4$ depends on assumptions about smoothness) \cite{bahri_explaining_2021}.

A related hypothesis is that the spectral density of the data covariance falls off as a power law, and 
in \cite{maloney_solvable_2022}  Eq. \ref{eq:scaling} is derived for a random feature
model with this covariance.  This hypothesis follows from the low dimensional hypothesis but it is more general,
for example these authors argue that additional features derived from the data (as in nonlinear models such
as FFN's) generally have the same spectrum as the original data.  One can also try to relate Eq. \ref{eq:scaling}
and corrections to it to hypotheses about how tasks are learned \cite{michaud_quantization_2023}.

What does the scaling of the information theoretic quantity
Eq. \ref{eq:loss} have to do with performance on tasks requiring intelligence?
{\it A priori}, not much, but one way to motivate a focus on it is to draw an analogy with particle physics.  
In the 30's cosmic ray observations gave strong
hints of new physics at higher energies, but the interesting events
were too rare and uncontrolled to draw solid conclusions.
Thus physicists were motivated to build accelerators.  These are not that expensive when they fit on a tabletop, but
rapidly grow in size and cost.  How large does an accelerator need to be?
The right measure is not its size {\it per se} but
rather the energy of the particles it can produce.  The physics relating size and energy is not trivial (due to
effects such as synchrotron radiation) but can be worked out, so one can make a good prediction of energy
reach.  Still, as one increases energy, will one find a smooth extrapolation of what came before, or will one
discover qualitatively new phenomena?  In the golden age of accelerator physics (the 50's-70's) much new
physics was discovered, mostly associated with new particles which are produced only above sharp energy thresholds.
Currently the highest energy accelerator is the Large Hadron Collider at Cern, where the Higgs particle was discovered
in 2012.  While we are still waiting for further important discoveries, the potential for discovery is
determined by measurable properties of the accelerator -- by energy and secondarily by intensity or ``luminosity'' --
which we can judge even in the absence of qualitative discoveries.  In the analogy, perplexity is playing a similar
role as an objective measure of language model 
performance defined independently of the more interesting qualitative behaviors which reflect ``intelligence.''

How far can one push this analogy?  Could perplexity be as central to language as energy is to physics?
Eq. \ref{eq:loss} has a fairly objective definition, so
the idea is not completely crazy.  But, not only was its relation to performance
on actual tasks not predictable in advance,  even after the fact clear ``thresholds'' or other signals for emergence
of tasks have not yet been identified \cite{wei_emergent_2022}.  Perhaps if there are universal thresholds,
evidence for them could be seen in humans.\footnote{ Thanks to Misha Tsodyks for this suggestion. }
More likely, additional variables (the quality and nature of the training corpus, details of the tasks, {\it etc.})
would need to be controlled to see them.  This is another question probably better studied in simpler tasks using synthetic
data.

The final topic we discuss is the behavior of the objective function (Eq. \ref{eq:loss})
as a function of training time.\footnote{This is roughly the time in which the gradient descent operates,
see Eq. \ref{eq:graddescent}.  In LLMs one often considers each data item only once in a training run, so it is related to
(but different from) dataset size.
}
In almost all ML runs, such a plot shows long plateaus interspersed with steep drops.
This has been interpreted in many ways, ranging from evidence about the nature of learning,
to a simple consequence of randomness of eigenvalues of the Hessian of the loss function.
A more recent observation is to compare training and testing accuracy on the same plot.
In \cite{barak_hidden_2022,power_grokking_2022} it was argued that these two metrics
improve at two distinct stages of training.  First, the model
memorizes training examples.  Later, it generalizes to the testing examples.
This ``grokking'' phenomenon
has been suggested as evidence for learning of circuits \cite{nanda_progress_2023}, an idea
we discuss in \S \ref{s:study}.

\section{ Simpler language models }
\label{s:details}

Here we describe a few generative language models in detail to fix the concepts.
As points of notation, let $\cal W$ be the set of words (or, if the reader prefers, numbers which index
a position in a list of words).  We denote the cardinality of a set $\cal S$ as $|\cal S|$,
so $|\cal W|$ is the number of distinct words.  The space of $N$-component real vectors is denoted
${\mathbb R}^N$.

The simplest model is the N-gram model defined in terms of the conditional probabilities
\be
P(w_{N} | w_1\; w_{2} \; \ldots \; w_{N-1} ) ,
\ee
which are all taken to be independent.  Given this minimalist assumption, a plausible way to
estimate them from the corpus is
\be\label{eq:Ngrammodel}
P(w_{N} | w_1\; w_{2} \; \ldots \; w_{N-1} ) =
\frac{ \mbox{Number of occurrences of } w_1\; w_2\;\ldots\; w_{N-1} \;w_{N}  }%
{ \sum_w \mbox{Number of occurrences of } w_1\; w_2\;\ldots\; w_{N-1}\; w } .
\ee
This simple model with $N=3$ or $4$ works better than one might think (see examples in \cite{jurafsky2009speech}) and
can be improved a bit by simple statistical tricks (``smoothing'').  But the exponential growth of the number
of strings in $N$ means that there is no hope of taking $N$ large enough to model even a single paragraph.  
The entire internet contains (in order of magnitude)
 $10^{12}$ words, and such a corpus will contain only a vanishingly small fraction of the 
likely twenty word strings.\footnote{
Statistical estimates of perplexity are in the 100's, and the best current LLMs have perplexity $\sim 20$.}

A more general principle which we can take from the N-gram model is the distributional hypothesis, which has been
pithily summarized as ``you shall know a word by the company it keeps.'' \cite{firth1957studies}
In other words, by proper use of the statistics of neighboring words, one can define quantities which
capture properties and even the meanings of words.  The simplest expression of this idea is the
co-occurrence matrix.  Before explaining this, let us mention a detail of practical systems, which
in place of words use ``tokens,'' meaningful components of words.
A physics illustration is the word ``supersymmetrization.''  Even for a non-physicist reader encountering
it for the first time, this word naturally breaks up into ``super,'' ``symmetry'' and ``ization,''  pieces which 
appear in many words and which are called tokens.  And not only does this decomposition apply to many words,
it helps to understand their meaning.  This process of replacing single words by strings of tokens (``tokenization'') 
is a first step in LLM processing, and henceforth when we say
``word'' we will mean word or token in this sense.

Given a corpus, we define its $N$-gram co-occurrence matrix $M_N$ to be the
$|\cal W|\times|\cal W|$ matrix whose $(w,w')$ entry counts the number of $N$-grams 
in the corpus containing both words.  This matrix defines a map from words to vectors
\be\label{eq:defiota}
\iota : {\cal W} \rightarrow {\mathbb R}^p
\ee
(where the dimension $p=|\cal W|$), by taking a word to the corresponding column of $M_N$.
Such a map is called a word embedding.

Applying this map to each word independently, we can map a string of $k$ words (in ${\cal W}^k$) 
to a string of vectors, and this is the next step (after tokenization) of LLM processing.  One might
worry that these are very high dimensional vectors with many zero entries, which seems wasteful.
A standard statistical cure for this problem is to do principal component analysis (PCA).
In words, instead of columns of $M_N$ we
use the columns of a $p\times |\cal W|$ matrix $Z$ chosen such that $Z^t Z$ is the
best rank $p$ approximation to $M_N$ in the sense that it minimizes $\mbox{tr}\, (Z^t Z-M_N)^2$.  One can do
better, but this gives the right idea.  

Next, we feed this string of vectors into some
machine learning model to get an output which we use to predict the
next word.  If we just want the most likely next word, a good way is to output a vector $v\in \IR^p$, and
choose the word $w$ which maximizes
the inner product $v \cdot \iota(w)$.  We denote this relationship as $v \sim \iota(w)$.  More generally,
the standard inverse map from a vector to a probability distribution on words
is the Boltzmann distribution on the inner products.  Explicitly, we postulate an inverse temperature $\beta=1/T$
and take\footnote{
$T$ is the temperature parameter which can be set in (say) the GPT user interface.  Also,
this ratio of exponentials is usually called ``softmax'' in machine learning as its $\beta\rightarrow \infty$
limit is the ``argmax'' function producing a vector whose nonzero components have the same index values
as the largest of the input(s).
}
\be\label{eq:softmax}
v \rightarrow P(w) = \frac{e^{\beta v \cdot \iota(w)}}{\sum_{w'} e^{\beta v \cdot \iota(w')}}
\ee

Here is an observation \cite{mikolov_efficient_2013} which supports the idea that word embeddings contain information
about meaning.  Since the embeddings are vectors, they can be added.  Consider the following equation:
\be\label{eq:embedrel}
\iota(\mbox{king}) - \iota(\mbox{man}) + \iota(\mbox{woman}) \sim \iota(?)
\ee
One might hope that the word which maximizes this inner product is ``queen,'' and indeed it is so.
There are many more such examples; empirically one needs the dimension $p\gtrsim 100$ for this
to work.  One can argue \cite{pennington_glove_2014,arora_latent_2019}
that it follows from relations between co-occurence statistics:\footnote{
Here $\#(w)$ denotes the number of occurences of ``$w$'' in the corpus.
These ratios can also be expressed in terms of the pairwise mutual information, $P_N(w,u)/P(w)P(u)$.
}
\be
\forall w,\, \frac{M_N(w,\mbox{king})/\#(\mbox{king})}{M_N(w,\mbox{queen})/\#(\mbox{queen}) }
\approx \frac{M_N(w,\mbox{man})/\#(\mbox{man})}{M_N(w,\mbox{woman})/\#(\mbox{woman}) }
\ee

Given these ideas and a map $F$ from a list of vectors to a vector,
we can now propose a very general class of $L$-gram autoregressive language models 
as the combination of the following steps:
\begin{enumerate}
\item Map the $L$ input words $w_i$ to $L$ vectors $\iota(w_i)$.
\item Apply $F$ to the list of these vectors to get a prediction vector $v$.
\item Use the inverse map Eq. \ref{eq:softmax} to get a probability distribution over words.
\end{enumerate}
Furthermore, if the map $F$ has parameters, given a corpus we can determine them  
by optimizing the function Eq. \ref{eq:loss} with respect to the parameters.
And once we bring in optimization, we can also optimize with respect to the coefficients of the 
embedding map Eq. \ref{eq:defiota}, so that we can dispense with co-occurence statistics.

This is the general prescription followed by the LLMs, and to complete it we just need to specify
a family of maps $F$.  One possibility
is to use a general (fully connected) feed forward neural
network (FFN, also called MLP for multilayer perceptron).  
We recall that an FFN is a composition of two general types of
functions, linear maps $W_i$ and nonlinear maps $\theta$, so that
\be\label{eq:ffn}
F(v) = W_d \circ \theta \circ W_{d-1} \circ \theta \circ \ldots \circ  W_1 \circ \theta \circ W_0.
\ee
In more concrete terms, the maps $W_i$ are multiplication by rectangular matrices of parameters
(usually called ``weights'' in this context), while the maps $\theta$ act independently on each
component of their input vector by a fixed nonlinear function such as $\tanh$ or (more typically)
ReLU (identity for $x\ge 0$ and zero for $x<0$).  The main fact we recall about FFN's
is that, in the limit that the number of parameters becomes large, they can approximate
any given function arbitrarily well \cite{Cybenko1989ApproximationBS}.
We refer the reader interested in learning more to \cite{devore_neural_2020,roberts_principles_2021}.

We can get a very natural deep learning version of the $L$-gram models
by using an FFN for the map $F$ in the prescription above \cite{bengio2000neural}.
Since this asked for a map from a list of vectors to a vector, we need to convert the
input list into a single vector.  This is easy: we can take the 
direct sum of the input vectors, {\it i.e.} the dimension $L\times p$ vector whose
components are the concatenated lists of their components.  Using today's FFNs, one could implement
this with $L \sim 100$ or so.  There does not seem to be much work on large fully connected
FFN language models, because by the time the technology advanced to this point the far more
efficient transformer models had taken over.  Still, they illustrate the general idea and also one of its
most obvious limitations.  Even with $L\sim 100$, often
predicting the next word requires remembering words which appeared farther back.
To solve this problem we need to incorporate some sort of memory into the model.

The simplest memory is an additional state variable which is updated with each word and used like
the other inputs.  To do this, we should take the state to be a vector in some ${\mathbb R}^q$.
This brings us to the recurrent neural network or RNN.  Its definition is hardly any more complex
than what we saw before.  With each word position (say with index $i$) we will associate a state
vector $s_i$ which can depend on words up to $w_i$ and on the immediately previous state.  Then,
we let the map $F$ determine both the next word and the next state as
\bea\label{eq:rnn}
(v_{i+1},s_{i+1}) = F( s_i, v_i, v_{i-1}, v_{i-2}, \ldots, v_{i-k+1} ),
\eea
where the parenthesis notation on the left hand side means that the output vector of $F$ is the
concatenation of two direct summand output vectors.  

Mathematically, Eq. \ref{eq:rnn} is a discrete dynamical system.
If we grant that $F$ can be an arbitrary map, this is a very general
class of systems.  One way of characterizing its generality is through
computational complexity theory, by asking what classes of computation
it can perform.  In \cite{siegelmann1992computational} it was argued that the
RNN is a universal computer, but this granted that the computation of $F$ in
Eq. \ref{eq:ffn} could use infinite precision numbers.  Under realistic assumptions
the right complexity class is a finite state machine, which can recognize
regular languages \cite{chen_recurrent_2018,weiss_practical_2018}.  We will
say more from this point of view in \S \ref{s:study}.

There are many variations on the RNN such as LSTM's \cite{hochreiter1997long}, each with their own advantages, but we must move on.

\section{ Recipe for an LLM }
\label{s:trans}

We are now ready to define the transformer model.\footnote{
Other reviews explaining these definitions include \cite{phuong_formal_2022,turner_introduction_2023}.}
It is simply another class of maps $F$ from 
lists of vectors to a vector to be used in the prescription above.  Indeed, it is a natural generalization of the
FFN which is associated to permutational symmetry.  This is in direct analogy to the
use of convolutional neural networks (CNNs) for image recognition, which are FFNs which are equivariant
under the symmetry of translations in two dimensions which is natural for the set of images.

A transformer is a composition of two types of functions (layers) taken in alternation, each mapping
an input list of $L$ vectors $\{ u_i \}$ to an output list of $L$ vectors $\{ v_i \}$.  One of these is an FFN
as previously discussed, but now applied to each embedding vector independently, so $v_i = F_{FFN}(u_i)$.

The other layer type is called attention, and it is defined as follows:
\bea\label{eq:attn}
\{ u_i \} &\rightarrow & \{ v_i = W \sum_{j=1}^i c_{i,j} u_j \} \\
c_{i,j} &\equiv & \frac{ \exp u_i \cdot B \cdot u_j }{ \sum_{j=1}^i \exp u_i \cdot B \cdot u_j }
\label{eq:attn2}
\eea
where $B$ is a learnable matrix whose elements are model parameters (equivalently, $u\cdot B\cdot v$ is a bilinear form)
and $W$ is a linear map (also learnable).\footnote{
One usually writes $B$ as the product of two ``key'' and ``query'' matrices,
and this can be used to restrict its rank. }

In words, an item $v_i$ in the output vector is (a linear transformation of)
a weighted sum of the inputs $u_j$ with $j\le i$ and can depend on any of them.\footnote{
The restriction  $j\le i$ to previous or current inputs is done to get an autoregressive model; one
can relax this for other purposes. }
The weights $c_{i,j}$ are given by a ``softmax'' or Boltzmann weight just as in Eq. \ref{eq:softmax}.
Thus there is a very general learnable way for each output to choose which of the input
vectors are most useful as inputs.   Suppose the product $u\cdot B\cdot v$ is the dot product,
then attention selects the input components $u_j$ most similar to the current unit's input, $u_j \sim u_i$
in the notation of \S \ref{s:details}.  The matrix $B$ allows for comparing different parts of the embeddings
and ignoring other parts,
in a way determined by optimizing the objective function Eq. \ref{eq:loss}.

Composing these two types of functions (or layers) produces a map from ${\mathbb R}^{p\times L}$
to ${\mathbb R}^{p\times L}$.  Often one takes, instead of the pure FFN and attention functions, sums
of these with the identity function (residual connections).  The FFNs generally have a single hidden layer
which can be of a different dimension, call this $p_h$.\footnote{
Explicitly, $v_i = W_1\cdot \mbox{max}(0,W_0 \cdot u_i + b_0) + b_1$,
where $b_{0,1}$ are more learnable parameters. }
Finally, while the language model prescription asked for a map to ${\mathbb R}^{p}$,
this is easily obtained by just taking the last vector in the final output list.

There are two more essential details to cover (and many minor details we will skip).
The first is the concept of ``attention head.''  The definition Eq. \ref{eq:attn} allowed for a general
linear transformation $W$ whose range is the output vector.  We are free to choose its dimension, call
it $q$, and typically one takes this to be much less than the embedding dimension $p$.  In return
one can use many copies of Eqs. \ref{eq:attn},\ref{eq:attn2} in parallel with different choices for $B$ and $W$, to produce
many outputs.  One can then concatenate these outputs to get a final output of dimension $p$.
These copies are called attention heads and we will denote their number by $H$, so $p=Hq$.

The second essential detail is that, so far, there is nothing in the definition that keeps track of the
order of the list of input vectors; the output of Eq. \ref{eq:attn} will be invariant under a general
permutation of the input vectors.  While this is an elegant property, it is
not what we want for processing language, for which the order of the words matters.
The cure for this is very simple: one takes as inputs not the word embeddings Eq. \ref{eq:defiota},
but the direct sum (concatenation) of these with positional embedding vectors, {\it i.e.}
vectors which encode the position (index) of the word in the string.  These can be
a combination of sines and cosines of various frequencies, such as \cite{vaswani_attention_2017}
\def\fff{  \frac{ \mbox{position} }{ 10000^{2i/d_{pos}} } }
\be
(e_{2i-1},e_{2i}) = (\,\cos\fff, \;\sin \fff\,); \qquad i \in \{1,\ldots,\frac{d_{pos}}{2} \}
\ee
One could instead treat these vectors as learnable parameters.  Still,
the trig function basis for positions
may be significant.  It has been generalized to represent other graph structures by using eigenfunctions
of the graph Laplacian as positional embeddings.  

The invariance of the transformer model under permutation symmetry is reminiscent of the point
we mentioned earlier, that translation symmetry motivates the CNN.  However
permutation symmetry is badly broken in language, even in the simplest formal languages,\footnote{
Compare the logical implications $A\rightarrow B$ and $B\rightarrow A$.}
and it is not obvious why this should be a useful property for the model to have.
One might argue that although any particular language breaks permutation symmetry, it acts naturally
on the ensemble of languages and thus should have a simple representation.  For example, besides
the usual infix arithmetic notation ``$a+b$'', one could instead use prefix ``$+\;a\;b$'' or postfix 
``$a\;b\;+$''.  Translating between these notations is arguably easier for
permutation invariant maps using position embeddings.
An opposing view would be that 
permutation symmetry is just a secondary property of the simplest model using attention, and that
the main point is to explain the value of attention.  In addition to its ability to select similar items, it
provides a simple way to take products of embedding vectors.  In computational complexity terms,
attention enlarges the class of circuits which can be simulated by a constant depth transformer
\cite{chiang_tighter_2023,edelman_inductive_2021,merrill_parallelism_2023,merrill_saturated_2022}.
Physics analogies of Eqs. \ref{eq:attn},\ref{eq:attn2}, especially to the Hopfield model, may be important 
\cite{hoover_energy_2023,ramsauer_hopfield_2021}.

A major practical advantage of the transformer over the RNN and other previous architectures is that the computations in the 
attention mechanism can be done in parallel, so (given sufficiently many processors) the time required
does not increase with the window length $L$.
This is by contrast with the RNN in which information
propagates from one word to the next, so a window of length $L$ requires time $L$ to process.
On the other hand the ability of each unit to pay attention to every previous unit means that the
total computation required by the transformer scales as $L^2$.  This is the limiting factor for increasing
$L$ and this is widely seen as a problem.  There has been a lot of work to improve this scaling, 
by removing some of the connections (as in sparse attention \cite{child_generating_2019}), 
by introducing multiscale structure,
or in other ways.

Let us summarize by listing the hyperparameters\footnote{
This term refers to model choices which are not learned through gradient descent.} 
and their values for the largest (175B) GPT-3 \cite{brown_language_2020}.
They are
\begin{itemize}
\item Embedding dimension $p=12288$ and hidden layer dimension $p_h=4p$.
\item Window length $L=4096$ or $8192$.
\item Depth $D=96$, counting both FFN (Eq. \ref{eq:ffn}) and attention (Eq. \ref{eq:attn}) layers.\footnote{
Some of the attention layers in GPT-3 are sparse.}
\item Number of heads $H=96$ (the equality with $D$ is a coincidence as far as I know). 
\end{itemize}
The total number of parameters is roughly $12Dp^2$.

As mentioned earlier, all of these parameters, and the parameters of the embedding map Eq. \ref{eq:defiota},
are determined as follows.  One generally starts with ``random'' initial conditions, usually meaning that
each parameter is drawn from a normal distribution with mean zero and variance chosen so that the linear
maps have expected norm independent of the hyperparameters.   As in random matrix theory, this typically
means $\mbox{var}(W_{i,j})\sim 1/p$, though there are refinements \cite{yang_tensor_2022}.  One then sequences
through the training corpus and performs a step of gradient descent of Eq. \ref{eq:loss} for each ``batch''
of words
(here a group of $\sim 10^6$ words).  In each step, the parameters $\vec\theta$ are modified as
\be\label{eq:graddescent}
\vec\theta \rightarrow \vec\theta - \eta \frac{\partial \CL_b}{\partial \vec\theta}
\ee
where $\CL_b$ is Eq. \ref{eq:loss} restricted to the batch, the conditional probability $P$ comes out
of Eq. \ref{eq:softmax} applied to the output of the transformer,
and $\eta$ is a positive real number  (the learning rate hyperparameter, here around $10^{-4}$).

The result of following this procedure on a dataset of natural language text,\footnote{
As always in ML it is important that the dataset be ``clean'' -- consistently
tokenized, not having too much garbage text or repetitions, {\it etc.}.  Many later LLMs also use
programming language code in the dataset.  Besides making code generation possible,
it has been reported that this improves performance
on natural language reasoning tasks. }
supplemented by many enhancements which are described
in the literature and in the model source codes but which may be less important for conceptual understanding,
is an LLM with the capabiliities we described.

\section{ Studying the internal workings }
\label{s:study}

The success of this procedure raises many questions.  Some can be asked about more or less any ML
model -- for example, questions about when and how optimization of the objective function Eq. \ref{eq:loss}
achieves ``good'' local minima (value near the global minimum and models which generalize well),
and the origin of scaling laws like Eq. \ref{eq:scaling}.  These are the subject of the general theory of
machine learning, for which we refer to \cite{bishop2006pattern,mezard2009information,roberts_principles_2021}
and much other work.

Other questions, and understanding the many striking abilities discussed earlier,
sound more specific to LLMs.  What would it mean to understand how ChatGPT writes poetry based on prompts,
or solves physics word problems?
At present this is by no means clear and it may be that entirely new concepts are needed to do this.
Still, I share the belief that we can go very far towards understanding LLMs by  
building on previous work in computer science, machine learning and AI, and many other fields.  
There is a well established field of statistical physics and ML \cite{mezard2009information}
which will surely contribute.  Physics ideas are also very relevant for tasks with spatial symmetry,
such as image generation \cite{sohl2015deep} and recognition \cite{cohen2016group}.
The unexpected mathematical simplicity of the transformer model means that mathematical insights
could be valuable.
We can also follow approaches used in neuroscience, psychology, and cognitive science.

An evident observation is that the paradigm of neuroscience -- careful study of the microscopic workings
of the system, following a reductionist philosophy -- is far more practical for ML models than it is
for human brains, as the microscopic workings are fully explicit.  This is not to say that it is easy, as we still
face the difficulty of extracting meaning from a system with billions of components and parameters.  How could 
we do this for LLMs?

One familiar starting point in neuroscience is to measure the activity of neurons and try to correlate it
with properties of the system inputs or outputs.  The ``grandmother cell'' which fires when a subject
sees his or her grandmother is an extreme (and controversial) example.  Better established are the ``place cells''
in the hippocampus which fire when an animal passes through a specific part of its environment.

Generally there is no reason why the representation should be so direct; 
there might be some ``neural code'' which maps stimuli onto specific combinations or patterns of activity.
The details of the neural code could even be different between one individual and the next.
Analogous concepts in LLMs are the maps from input strings to intermediate results or ``activations.''
The first of these is the embedding map Eq. \ref{eq:defiota}.  Considering each layer in succession,
its outputs (sometimes called ``contextualized embeddings'') also define such
a map.   The details of these maps depend on details of the model, the training dataset 
and the choices made in the training procedure.  Besides the hyperparameters, these include 
the random initializations of the 
parameters, the order in which data items are considered in training and their grouping into batches.
Even small differences can be amplified by the nonlinear nature of the
loss landscape.

One way to deal with this indeterminacy is to look for structure in the maps which does not depend on these choices.
The linear relations Eq. \ref{eq:embedrel} between word embeddings are a very elegant example,
telling us (and presumably the model) something about the meanings of the words they represent.  Moving on
to the later layers, one can ask whether contextualized embeddings carry information about the grammatical role of a word, about other words it is associated to (such
as the referent of a pronoun), {\it etc.}.  One can go on to ask whether any of the many structures which -- one
would think -- need to be represented to understand the real world, are visible in these embeddings.

Many structures are too intricate to show up in linear relations.  A more general approach is to
postulate a ``target'' for each training data item and train a ``probe'' model (usually an FFN) to predict it from the
embeddings.  If this works, one can go on to modify the internal representation 
in a minimal way which changes the probe prediction, and check if
this leads to the corresponding effects on the output 
(see \cite{Belinkov2021ProbingCP} and references there).

This procedure is simpler to explain in an example.
A pretty example of probing for a world model is the recent work of Li {\it et al} 
\cite{li_emergent_2023} (see also \cite{toshniwal_chess_2022})
on representations in a transformer model trained to play the board game Othello.\footnote{
For readers not familiar with this game, two players alternate in placing black and white
tiles on an $8\times 8$ board, and each move results in ``flipping'' some opponent pieces
to the player's color.  The main point for us is that the function from moves to board state is
easily computable yet very nonlocal and nonlinear. 
}
They train a model ``Othello-GPT''
\footnote{
While this model shares the GPT architecture, it is not
trained on any language data, just on Othello games.
} 
to take as input a sequence of 60 legal moves,
for example ``E3 D3 ...'' in the standard algebraic notation, 
and at each step to predict the next move.  The trained model
outputs only legal moves with very high accuracy, and the question is whether
this is done using internal representations which reflect the state of the game board, say the presence
of a given color tile in a given position.
Following the probe paradigm, they obtain FFNs which, given intermediate activations, can predict whether
a board position is occupied and by which color tile.
Furthermore, after modifying the activations so that the FFN's output has flipped a tile color, 
the model predicts legal moves for the modified board state, confirming the identification.
Neuroscientists can only dream of doing such targeted experiments.

Numerous probe studies have been done on LLMs.
One very basic question is how they understand
grammatical roles and relations such as subject, object and the like.
This question can be sharpened to probing their internal representations for 
parse trees, a concept we review in the appendix.  To get the targets for the probe,
one can use a large dataset of sentences labeled
with parse trees, the Penn Treebank \cite{marcus1993building}.
This was done for BERT in 
\cite{chi_finding_2020,hewitt_structural_2019,manning2020emergent} by the following procedure:
denote the embedding (in a fixed layer) of word $i$ as $u^i$, then the model learns a
projection $P$ on this space, such that the distances $d(i,j)\equiv ||P(u^i-u^j)||$
in this inner product well approximate the distance between words $i$ and $j$ defined as
the length of the shortest path connecting them in the parse tree.
For BERT (with $d\sim 1000$) this worked well with 
a projection $P$ of rank $\sim 50$.

Once one knows something about how information is represented by the models, one can go on to try
to understand how the computations are done.  One approach, also analogous to neuroscience, is to look
for specific ``circuits'' which perform specific computations.  An example of
a circuit which appears in trained
transformer models is the induction head \cite{elhage_toy_2022,olsson_-context_2022}.  
This performs the
following task: given a sequence such as ``$A\, B\,\ldots\, A$'' it predicts a repetition, in this example ``$B$.''
The matching between the tokens (the two $A$'s in the example) is done by attention.
A number of works have proposed and studied such circuits, with various motivations and using various
theoretical lenses:  interpretability and LLMs \cite{olah_circuits_2022},
in-context learning \cite{olsson_-context_2022,akyurek_what_2022}, 
formal language theory \cite{merrill_linguistic_2020,chiang_tighter_2023},
computational complexity theory \cite{edelman_inductive_2021,liu_transformers_2022}, 
{\it etc.}.

Reverse engineering a large network {\it ab initio}, {\it i.e.} with minimal assumptions about what it is doing,
seems challenging, but maybe automated methods will be developed \cite{chughtai_toy_2023,friedman_learning_2023}.
Another approach is to first develop a detailed computational model (CM) to perform a task without
looking too much at the system under study, and then
look for evidence for or against the hypothesis that the system under study uses it.
This approach also has a long history in neuroscience \cite{marr2010vision} and ways to
test such hypotheses have been much discussed.  As an example of a
research tactic which does not require opening the black box,
one can consider illusions which fool the system in some way.  The response
to these will often depend on contingent and non-optimal aspects of the model, so one can
distinguish different models which solve the same task.  A new class of predictions which becomes
testable for LLMs is to look at performance as a function of 
model size (depth; number of parameters).
A particular CM might require a certain model size or dataset properties in order to perform well.
And of course, one can open the black box:
by assuming a particular CM,
one can make predictions for what probe experiments should work.

Simple tasks studied in this approach include modular addition \cite{nanda_progress_2023}
and linear regression \cite{akyurek_what_2022}, where several CM's (gradient descent, ridge
regression and exact least squares) were compared.  Turning to language processing, 
a CM for parsing by transformer LLMs was developed in Zhou {\it et al} \cite{zhao_transformers_2023}.
While this is too lengthy to explain in detail here, let us give the basic idea, starting from the PCFG 
framework discussed in the appendix.  Rather than try to represent a parse tree in terms of nodes and
edges, it is represented by giving each position $i$ in the list of words a set of variables $\alpha_{i,t,j}$,
where $t$ indexes a nonterminal (a left hand side of a rule) and $j$ is another position.
If $\alpha_{i,t,j}$ is turned on, this means that a rule with $t$ on the l.h.s. was used to generate that part
of the tree stretching from position $i$ to position $j$.  This can be generalized to let $\alpha_{i,t,j}$ be the
probability that a rule is used.  These variables (and additional variables $\beta$ describing the rules used
higher in the tree)
satisfy simple recursion relations (the Inside-Outside parsing algorithm \cite{manning1999foundations}).  
If the rules have at most two symbols on the r.h.s.,\footnote{ 
One can rewrite any grammar to have this property (Chomsky normal form) by introducing more nonterminals.
}
these recursion relations
are quadratic in the variables.  By encoding
the $\alpha$ variables as components of the embedding, they can be implemented using attention.

Naively, this model predicts that embedding dimension $p$ must be very large, of order the number of
nonterminals times the length of a sentence.  Since realistic grammars for English have many hundreds
of nonterminals, this seems to contradict the good performance of transformers with $p \sim 1000$.
This problem is resolved by two observations, of which the first is that one can get fairly good parsing
with many fewer ($\sim 20$) nonterminals.  The second is compression, that embeddings and circuits
which are simple and interpretable can be mapped into more ``random-looking'' lower dimensional 
forms.   This is a well understood concept for metric spaces \cite{matousek_lecture_2013}, which was 
implicit in the discussion of word embeddings in \S \ref{s:details}.  There the simplest construction 
(the co-occurence matrix) produced vectors with one component for each word, but by projecting on
a subspace one could greatly reduce this dimension with little loss in accuracy.  The generalization of
these ideas to neural networks seems important.

Once one believes an LLM is carrying out a task using a particular circuit or CM, one can go on to ask
how it learned this implementation from the data.  One can get theoretical results in the limit of
infinite training data and/or for simple tasks in which the dataset is constructed by a random process.
Learning in transformer models trained on realistic amounts of data 
is mostly studied empirically and using synthetic data.  A few recent interesting 
works are \cite{akyurek_what_2022,gromov_grokking_2023,nanda_progress_2023}.  
Intuitively one expects that simpler instances of a task are learned first, allowing
the model to learn features which are needed to analyze more complex instances, and there is a lot
of evidence for this.  The idea that many submodels can be learned simultaneously, including straight
memorization and submodels which rely on structure, also seems important.
Ultimately learnability is crucial but we should keep in mind that in analogous questions in physics,
evolution, and so on, it is much easier to understand optimal and critical points in the landscape than
to understand dynamics.

This brings us to in-context learning, the ability of an LLM to perform diverse tasks
given only a few examples of input-output pairs.  The simplest hypothesis is that the model
has learned the individual tasks, and the examples are selecting a particular task from this repertoire.
It has been argued that this is guaranteed to happen (in the infinite data limit)
for a model trained on a mixture of tasks \cite{xie_explanation_2022,wies_learnability_2023}.
If the many tasks have common aspects (for example parsing might be used in any linguistic task),
one can ask how the model takes advantage of this, a question discussed in \cite{hahn_theory_2023}.
  
Understanding LLMs is a very active research area and there is much more we could say, but 
let us finish by summarizing the two main approaches we described.
One can postulate a representation and a computation designed to perform a task,
and look for evidence that the LLM actually uses the postulated structure.  Alternatively,
one can look for a function in some simpler class (such as digital circuits) which well approximates the function
computed by the transformer model, and then ``reverse engineer'' the simpler function to find out what it is doing.
Either or both of these procedures could lead to interpretable systems and if so, are answers to the question ``what
has the LLM learned.''  There is no guarantee that they will work and it might turn out
that one cannot understand LLMs without new ideas, but they deserve to be tried.

\section{ Questions and discussion }
\label{s:how}
\label{s:end}

Large language models have revolutionized computational linguistics and opened up many
new applications of AI.  Understanding how they work is both straightforward (we explained it in \S \ref{s:trans})
and at the same time an outstanding scientific challenge.  This is because the question
``how do they work'' has multiple meanings.  On the one hand, LLMs are a relatively simple
solution to the task of predicting the likely next word in a text.  On the other hand, they also seem to
perform many other tasks which require intelligence, such as solving the physics word problem in Figure \ref{fig:minerva}.
While we do not have a strong understanding of what a system which can perform these tasks must do, 
a vast body of work in cognitive science and AI supports one's first naive intuition that such a system
must be doing sophisticated analyses of language, must contain models of the real world, and must be
able to do fairly general logical reasoning.  Before it was demonstrated,
the idea that all this could be learned as a byproduct of
word prediction would have seemed hopelessly optimistic, had anyone dared to suggest it. 

Extraordinary claims should be greeted with skepticism.  One must guard against
the possibility that a successful ML system
is actually picking up on superficial aspects or statistical regularities of the inputs, the ``clever
Hans'' effect.   Addressing this is an important function of the benchmark evaluations discussed in \S \ref{s:pheno}.
Of course as LLMs get good at performing tasks of practical value, the skeptical position becomes hard to maintain.

Intelligence and language are incredibly complex and diverse.  According to Minsky,\footnote{
What magical trick makes us intelligent? The trick is that there is no trick. The power of intelligence stems from our vast diversity, not from any single, perfect principle. \cite{minsky1988society}}
this diversity is a defining feature of intelligence.
The goal of understanding LLMs (or any general AI) will not be accomplished
by understanding all of the content in their training data, the ``entire internet.''   
Rather, the trick we need to understand is how a single system can learn from this diverse corpus to
perform a wide range of tasks.  Theories of ``what is learnable'' are a central part of computer science
\cite{kearns1994introduction}.  Although theoretical understanding has a long way to go to catch up with
LLM capabilities, for simpler and better understood tasks much is known.

In these notes we mostly looked at this question through the lens of computer science, and took as
the gold standard for explaining how an LLM learns and performs a task, a computational model expressed
as an algorithm or a circuit together with arguments that the trained LLM realizes this model.
This point of view has many more insights to offer, but before we discuss them let us consider some other
points of view.  In \S \ref{s:study} we drew the analogy between detailed study of
transformer circuits and neuroscience -- what others can we consider?

Another analogy is with cognitive psychology.  LLMs are sufficiently human-like to make this interesting, 
and there is a growing literature which applies tests and experimental protocols from psychology 
to LLMs, see for example \cite{hagendorff_machine_2023} and the many references there.
When discussing this, we should keep in mind
the vast differences between how humans and LLMs function.
Human brains are not believed to use the backpropagation learning algorithm, indeed
it has been argued that biological neural systems cannot use it  \cite{Crick1989TheRE}. 
Perhaps related to this, brains are not feed-forward networks but have
many bidirectional connections.  Whatever brains are doing, it works very well:
LLMs (like other current deep learning systems) need far more training data than humans.
Furthermore, the LLMs we discussed do not interact with the world.
Some argue that on philosophical grounds,
a model trained only on language prediction can never learn meaning \cite{bender-koller-2020-climbing}.
While I do not find this particular claim convincing, I agree that
we should not assume that LLMs perform tasks the same way humans do.
Still both similarities and differences are interesting;
can we make the analogies with cognitive psychology more precise?

One analogy \cite{bengio2019,goyal_inductive_2022},
is with the well known concept of ``fast and slow thinking'' in behavioral psychology
\cite{kahneman2011fast}.  To summarize, humans are postulated to have two modes of thought,
``system 1'' which makes fast, intuitive judgments, and ``system 2'' which can focus attention and do calculations,
logic, and planning.  While system 2 is more general and less error-prone, using it requires conscious attention and effort.
According to the analogy, LLMs implement system 1 thinking, and are weak at system 2 thinking.

In \cite{mahowald_dissociating_2023} it is argued that LLMs have ``formal linguistic competence''
but not ``functional competence.''  In plainer terms, they are solving problems by manipulating language
using rules, but they lack other mechanisms of human thought.  
While it may be surprising that a purely rule-based system could do all that LLMs can do,   
we do not have a good intuition about what rule-based systems with billions of rules can do.

What are the other mechanisms?
There is a long-standing hypothesis in cognitive science, modularity of mind \cite{fodor1983modularity},
according to which the human brain has many ``mental modules'' with different capabilities.  These include
a language module of the sort that Chomsky famously advocated and many others, including one for geometric and physical
reasoning, another for social reasoning and theory of mind, and perhaps others.
Notably, formal logic and mathematical reasoning
seem to call upon different brain regions from those which specialize in language \cite{amalric_distinct_2019},
suggesting that these functions are performed by different mental modules.
One can thus hypothesize that LLMs have commonalities with the human language module and might
be useful scientific models for it,\footnote{
Chomsky rejects this idea, saying that
``The child’s operating system is completely different from that of a machine learning program.''
(New York Times, March 8, 2023).
}
but that progress towards human level capability will eventually stall without analogs of the other modules.
\cite{lake_building_2016}

A related claim is that current LLMs, even when they perform well on benchmarks,
do not construct models of the world.   Consider
reasoning about spatial relations -- for example if A is in front of B is in front of C, then A is in front of C.
Such reasoning is greatly facilitated by representing the locations of objects in space, perhaps in terms of
coordinates, perhaps using ``place cells'' or in some other non-linguistic way.
If distance from the observer is explicitly represented and used in reasoning,
then it becomes hard to get this type of question wrong.
Conversely, to the extent that LLMs do get it wrong, this might be evidence that they lack this type
of world model or cannot effectively use it.

There are many papers exhibiting LLM errors and suggesting such interpretations, but often one finds
that next years' model does not make the same errors.  At the
present rate of progress it seems premature to draw any strong conclusions.  My own opinion is that
there is no barrier in principle to LLMs constructing internal non-linguistic models of the world, and
the work \cite{li_emergent_2023} on Othello-GPT discussed in \S \ref{s:study} is a nice demonstration
of what is possible.  This is not to say that any and all models can be learned, but rather that it might be better
for now to focus on other significant differences between LLM and human
reasoning, of which there are many.  I will come back to this below.

If LLMs and other connectionist systems do not work in the same way as brains,
what other guidance do we have?  In  \S \ref{s:study} we discussed one answer,
the hypothesis that they work much like the algorithms and circuits studied
in computer science.  Perhaps trained LLMs implement
algorithms like those designed by computational linguists, or perhaps new
algorithms which were not previously thought up but which can be understood in
similar terms.  In either version this is still a hypothesis, 
but if we grant it we can draw on insights
from theoretical computer science which apply to all such algorithms.

Computational complexity theory
\cite{arora2009computational,wigderson2019mathematics}
makes many statements and conjectures about how the time and space
required by a particular computation depends on the size of the problem
(usually meaning the length of the input).  The most famous of these,
the $\P\ne\NP$ conjecture, states (very loosely) that for problems
which involve satisfying general logical statements, finding a solution
can be much harder than checking that the solution is correct. 

From this point of view, a central question is the complexity class of circuits
which can be realized by constant depth transformers, meaning that the number
of layers does not grow with the window size.  Roughly, this is the complexity
class $\TC^0$ of constant depth circuits with threshold gates
\cite{chiang_tighter_2023,edelman_inductive_2021,%
merrill_parallelism_2023,merrill_saturated_2022}.
Of course in an
autoregressive LLM one can repeat this operation to compute a sequence of
words: thus the circuit defines the transition function of a finite state machine (FSM) where
the state is the window, and the LLM has learned to simulate this FSM.
If a natural algorithm to perform a task
is in a more difficult complexity class than the FSM can handle, this is
a reason to think the task cannot be learned by this type of LLM.
Conversely, one might conjecture that any task for which there is an algorithm
in this class can be learned, at least in the limit of an infinite amount of training data.

What about the lenses of pure mathematics, theoretical physics and allied fields?  Besides my own
personal interest in them, these fields have made substantial contributions to 
statistics and machine learning,
especially the interface between statistical physics and machine learning is a vibrant field of research
\cite{krzakala2016statistical,mezard2009information}.
Spin glass theory made a very deep impact, starting with the Hopfield model
and developing into a far-reaching theory of optimization landscapes and
complexity.
Random matrix theory is central to high dimensional statistics 
\cite{johnstone_high_2006} and 
in many approaches to understanding deep learning \cite{roberts_principles_2021}.
Mathematical approaches to language such as
\cite{bradley_enriched_2021,coecke_mathematical_2010,manin_semantic_2016,marcolli_mathematical_2023}
can reveal new structure and provide deeper understanding.

Another reason to think pure mathematics and theoretical physics have more to contribute is that
neural networks, transformers, and many of the models of neuroscience,
are formulated in terms of real variables and continuous mathematics.
By contrast, computer science is largely based on discrete mathematics, 
appropriate for some but not all questions.
Perhaps word embeddings have important geometric properties,
or perhaps the dynamics of gradient descent are best understood through
the intuitions of continuous mathematics and physics. 
Arguments such as those in \S \ref{s:study} which reduce neural networks to digital
circuits, even if they do explain their functioning, may not be adequate to explain
how they are learned.

Having at least mentioned some of the many points of view, let me combine these insights and speculate a
bit on where this is going.  Let me focus on three capabilities which seem lacking in current LLMs:
planning, confidence judgments, and reflection.

Planning, solving problems whose solution requires choosing a series of actions and/or the consideration of
future actions by other agents, is one of the core problems of AI.  Making plans generally requires search, and in general search is hard (assuming $\P\ne\NP$).
A familiar example is a chess program, which searches through a game tree to judge the longer term
value of a candidate move by hypothesizing possible future moves. 
While much of the success of AlphaGo and AlphaZero is attributed to reinforcement learning by self-play,
they also search through game trees; indeed the Monte Carlo tree search algorithm on which they built
\cite{coulom2006efficient,browne_survey_2012}
was considered a key enabling breakthrough.

By contrast, LLMs have no component dedicated to search.   While it does not seem impossible
that search trees or other structures could be learned internally (like world models),
it seems intuitively clear 
that an autoregressive model which predicts one word at a time and cannot
go back to revise its predictions in light of what comes later will be seriously handicapped in planning.
This observation is motivating a fair amount of current work on ways to incorporate search.
LeCun has suggested adding a dynamic programming component to search through multiword
predictions, as part of his ``path towards autonomous machine intelligence'' \cite{LeCun_2022}.
Another proposal, the ``tree of thoughts'' model \cite{yao_tree_2023}, works with a search tree of LLM responses.
A system which uses hierarchical planning for mathematical theorem proving was developed in \cite{jiang_draft_2022}.

The next capability on my list, making and working with confidence judgments, has to do with the
well known ``hallucination'' problem, that LLMs often simply invent statements,
including untrue facts and imaginary citations.
While advantageous for a poetry generator, and bearable for a system which makes suggestions which an expert
human user will verify, this is a huge obstacle to many practical applications.
Thus it is the subject of a great
deal of research -- a few of this month's papers are 
\cite{feldman_trapping_2023,li_inference-time_2023,lightman_lets_2023}.
Perhaps by the time you read these words there will have already been major progress. 

Why are LLMs producing these hallucinations? 
One intuition is that they are doing some sort of compression, analogous
to JPEG image compression, which introduces errors \cite{ted_chiang}.  This point of view suggests that the problem will
eventually be solved with larger models and perhaps better training protocols which focus on the more
informative data items \cite{sorscher_beyond_2022}.  

A related intuition is that the problems follow from inability to properly generalize.  This comes
back to the point about ``world models'' -- a correct model, for example an internal encoding of place information,
by definition correctly treats the properties being modeled.  Now suppose we grant that the LLM is 
solving some class of problems, not by constructing such a model, but by rule-based reasoning.
In other words, the LLM somehow learns rules from the corpus which it uses to make 
particular inferences which agree with
the model.  While such rules can cover any number of cases,
there is no clear reason for such a rule set to ever cover all cases.

Another intuition is that the training data contains errors and this is reflected in the results.
Certainly the internet is not known for being a completely reliable source of truth.
This intuition also fits with the observation that adding code (computer programs) to the training set
improves natural language reasoning.  Code is a good source of rules because almost all of it has been 
debugged, leading to rules which are correct in their original context (of course they might not be correctly applied).
It is a longstanding question whether internal representations (both in AI and in humans) are shared between
different natural languages; it would be truly fascinating to know how much they are also shared with code.
If this intuition is right, then LLMs reasoning capability might be improved by training on far more code and
other content which is guaranteed to be correct.  Such content could be generated synthetically as
tautologies, or even better as formal verified mathematics (as proposed in \cite{Szegedy2020APP}).

Here is a different point of view:
the problem is not that the systems make things up, after all creativity
has value.  Rather, it is that they do not provide
much indication about the confidence to place in a particular output, and do not have
ways to adapt their reasoning to statements known at different levels of confidence.  Much of our reasoning involves uncertain claims and claims which turn out to be false, 
the point is to distinguish
these from justified claims and keep track of our confidence in each belief.
While it is possible to extract confidence scores from LLMs \cite{kadavath_language_2022},
there is also a philosophical point to make here:  not all facts have the same epistemological status.
Some facts are grounded in evidence; others are true by definition.

LLMs are of course statistical models.   Even for a completely deterministic task, say doing arithmetic, 
a statistical approach to learning is very powerful.  This is because learning based on inputs which consist
of finitely many training examples, given in a random order, is naturally formulated in statistical terms.
But without making additional non-statistical
assumptions, one can never go from almost 100\% confidence to 100\% confidence.

This difference is crucial in many aspects of human thought.  Of course, logical reasoning and mathematics
stand out as prime examples.  Long chains of reasoning are only possible if the individual links are reliable.
But it is also crucial in social reasoning. 
There is an essential difference between statistical and evidence-based statements,
say ``Michael is a popular name,'' and tautological, definitional and
descriptive statements such as ``My name is Michael.''
While the first statement might be a subject of discussion,
a model which can get confused about the second statement
is clearly missing a defining aspect of human thought,
and will lose the confidence of its interlocutor.
Perhaps epistemological status and tautological correctness need to be somehow represented in the model.  It need not
be designed in, but the model needs to be given additional signals beyond next word prediction to learn it.

The third point on my list, reflection, does not seem to be much discussed, but to me seems just
as important.  In computer science, reflection is the capability of a system to work with its programs
as a form of data \cite{smith1982procedural,wikipedia_reflection}.
This is naturally possible for a computer programmed
in assembly language, in which instructions are encoded in integers.  To some extent it is also possible 
in Lisp, in which programs are encoded in a universal list data structure.  As type systems and
other programming language refinements are introduced, reflection becomes more difficult to provide,
but it is necessary for systems-level programming and makes various standard tasks easier to implement.

Since an LLM operates on language, reflection for an LLM is the ability to work with its internal
model in linguistic terms.  This is related to ML interpretability, the ability to translate a model 
into understandable terms.  In \S \ref{s:study} we discussed interpretability of LLMs in terms of 
circuits and computational models, implicitly leaving these for a human to interpret and understand.
One can imagine an ``interpretation engine'' which given a model, automatically produces a more interpretable description,
in terms of circuits, rules, or even a description of the model's functioning in natural language.  
Given such an interpretation engine, by applying it to an LLM and sending its output as an input to the LLM, we can
implement a form of reflection. 

A basic human capability which corresponds to this process is the translation from procedural or other implicit forms of
memory to linguistic, explicit memory.  Very often, we learn by doing -- riding a bicycle, solving math
problems, interacting socially.  We then reflect on what we have learned -- in some unconscious way --
and occasionally come up with verbal observations, summaries, in a word reflections.  It is fascinating
that combining the ideas we discussed brings us into contact with such topics.

To conclude, and for what it is worth, out of the forty years I have followed AI,
this is by far the most exciting period.  I agree with those who think LLMs are a major
milestone and believe the ideas behind them -- including the transformer architecture -- will 
remain important even in the light of future progress.  The questions they raise are interesting and important enough
that -- even as the specialists make remarkable progress -- we need not leave the field to them, but as scientists
and thinkers we should engage and try to contribute.

\appendix

\section{ Grammars and parsing }
\label{s:gofai}

Most readers will have encountered the idea of ``sentence diagram,'' which graphically
represents the decomposition of a sentence into clauses with a subject, verb and object, the assignment
of adjectives and prepositional phrases to the nouns and verbs they modify, and so on.
Formal versions of this concept are foundational in linguistics and computer science,
and a short introduction (or review) is a good way to bring the general ideas we are discussing to life.

A formal grammar can be given by a set of ``production rules'' 
which can be used to generate grammatical strings.  A simple
example is in Figure 3. 
\def\EXPR{\mbox{EXPR}}
\def\TERM{\mbox{TERM}}
\def\VALUE{\mbox{VALUE}}
\def\NUMBER{\mbox{NUMBER}}
\def\VAR{\mbox{VAR}}
\begin{figure}[ht] \label{fig:grammar}
\small
\bea\label{eq:grammar}
\EXPR & \rightarrow & \TERM + \EXPR \\ \nonumber
\EXPR & \rightarrow & ( \,\EXPR \,) \\ \nonumber
\EXPR & \rightarrow & \TERM \\ \nonumber
\TERM & \rightarrow & \VALUE * \TERM \\ \nonumber
\TERM & \rightarrow & ( \,\EXPR\, ) \\ \nonumber
\TERM & \rightarrow & \VALUE \\ \nonumber
\VALUE & \rightarrow & x \\ \nonumber
\VALUE & \rightarrow & y \\ \nonumber
\VALUE & \rightarrow & 1 
\eea
\caption{A grammar for arithmetic expressions. }
\end{figure}
Each line is a rule, which is made up of two strings of symbols separated by $\rightarrow$,
the left hand side or lhs and right hand side or rhs.
These symbols can be ``terminals'' which appear in the language (such as 
$+$, $*$, $x$, $0$
and so on in our example) or ``nonterminals'' which do not (such as $\TERM$).

These rules are used as follows: We start with a string $S$ containing a distinguished ``start'' symbol (here \EXPR).
We then iterate the following process:  choose a rule whose lhs occurs in $S$, and apply it by substituting
one occurrence of this lhs in $S$ with the rhs.  Every string $S$ which can be obtained by a finite sequence of 
these operations is considered grammatical, and by keeping track of the rule applications we get a parse tree.
This is a graph whose nodes are symbols and whose edges connect a symbol which appears on the
lhs of a rule application with the nodes for each of the symbols which appear on the 
rhs.\footnote{
One can see examples for English sentences in the Wikipedia article ``Parse tree.''
}
A good exercise is to work out the parse tree for the expression $y + 1 * x$
and check that multiplication takes precedence over addition.

Our example of a grammar is a context-free grammar, 
meaning that the left hand side of each rule consists of a single symbol.
If we do not put this restriction, the resulting class of languages are universal computers (and thus suffer from
potential undecidability).  There is also a more restricted class of grammars called regular grammars (this
hierarchy was found by Chomsky), but these cannot describe nested structures such as the parentheses of 
 Eq. \ref{eq:grammar}.   The context-free grammars are the right degree of complexity for many
purposes.  In particular, programming languages and the formal languages of mathematical logic can be
described using CFG's
and thus the algorithms for working with them and associated theory are well developed.

Besides recognizing and parsing languages, one can describe other linguistic tasks in similar terms.
A trivial example would be word replacement, with rules such as $\mbox{OLD}_i \rightarrow \mbox{NEW}_i$.
Realistic tasks benefit from frameworks with more structure.  For example, to use the grammar in Eq.
\ref{eq:grammar} to do arithmetic, we would be much better off with a framework in which the token
\VALUE\ carries an associated numerical or symbolic value.
This can be done with the framework of attribute grammars.  When we
suggest in \S \ref{s:end} that LLMs perform natural language tasks using systems of large numbers of rules,
we have this sort of extended grammatical framework in mind.

CFG's are not really adequate for natural languages, with their inherent ambiguity
and their many special cases and exceptions.
A more general
formalism is the probabilistic CFG.  This is obtained by associating a probability distribution
to each symbol which appears on the left hand side of a rule (the nonterminals).  For example, we might
stipulate that a \VALUE\ has a 75\% chance to be a number and a 25\% chance to be a variable.  With
this information, a PCFG defines a probability distribution on strings, which gives zero probability to
nongrammatical strings.

A symbolic approach to parsing would propose two primary algorithms.  One is a parser,
which given a grammar and an input produces the parse tree.  
Another would be an algorithm for
learning a grammar from a corpus.  Since any finite corpus can be described by many grammars,
PCFG's are better suited than CFG's to this problem.  In any case the learning and parsing algorithms
are not necessarily related.

In the connectionist approach followed by LLMs, these two algorithms are subsumed into the definition of
a model which can parse any PCFG whose rules are encoded in its weights.  By training this on a corpus, the
model learns a particular PCFG which generates the corpus.  Interpretability as discussed in \S \ref{s:study}
then means reversing this relation, by extracting a parser and PCFG from the trained model.

\bibliography{trans,airefs}

\end{document}